\documentclass[onecolumn,english,12pt]{article}
\usepackage[T1]{fontenc}
\usepackage[latin1]{inputenc}
\usepackage{geometry}
\usepackage{array}
\usepackage{float}
\usepackage{amsmath}
\usepackage{graphicx}
\usepackage{amssymb}

\makeatletter

\floatstyle{ruled}
\newfloat{algorithm}{tbp}{loa}
\floatname{algorithm}{Algorithm}

\usepackage{times}

\usepackage{babel}
\makeatother
\begin{document}

\title{Video Segmentation via Diffusion Bases}
\author{Dina Dushnik$^1$~~Alon Schclar$^2$\footnote{Corresponding author: alonschc@mta.ac.il, Tel:+972-54-5456226, Fax:+972-3-6803342}~~Amir Averbuch$^1$\\
$^1$ School of Computer Science, \\ Tel Aviv University, Tel Aviv 69978, Israel\\
$^2$ School of Computer Science, \\ Academic College of Tel-Aviv Yafo, Tel Aviv 61083, Israel}

\date{}
\maketitle

\begin{abstract}
Identifying moving objects in a video sequence, which is produced by
a static camera, is a fundamental and critical task in many
computer-vision applications. A common approach performs background
subtraction, which identifies moving objects as the portion of a
video frame that differs significantly from a background model. A
good background subtraction algorithm has to be robust to changes in
the illumination and it should avoid detecting non-stationary
background objects such as moving leaves, rain, snow, and shadows.
In addition, the internal background model should quickly respond to
changes in background such as objects that start to move or stop.

We present a new algorithm for video segmentation that processes the
input video sequence as a 3D matrix where the third axis is the time
domain. Our approach identifies the background by reducing the input
dimension using the \emph{diffusion bases} methodology. Furthermore,
we describe an iterative method for extracting and deleting the
background. The algorithm has two versions and thus covers the
complete range of backgrounds: one for scenes with static
backgrounds and the other for scenes with dynamic (moving)
backgrounds.
\end{abstract}

\textbf{keywords}: Video Segmentation, Background subtraction, Markov processes, Graph algorithms.

\section{Introduction \label{sub:introduction}}

Video surveillance systems, tracking systems, statistical packages
that count people, games, etc. seek to automatically identify
people, objects, or events of interest in different environment
types. Typically, these systems consist of stationary cameras, that
are directed at offices, parking lots, playgrounds, fences and so
on, together with computer systems that process the video frames.
Human operators or other processing elements are notified about
salient events. There are many needs for automated surveillance
systems in commercial, law enforcement, and military applications.
%A continuous
%24-hour monitoring of surveillance video to alert security officers
%of a burglary in progress or a suspicious individual loitering in
%the parking lot is needed.
In addition to the obvious security applications, video surveillance
technology has been proposed to measure traffic flow, detect
accidents on highways, monitor pedestrian congestion in public
spaces, compile consumer demographics in shopping malls and
amusement parks, log routine maintenance tasks at nuclear
facilities, and count endangered species. The numerous military
applications include patrolling national borders, measuring the flow
of refugees in troubled areas, monitoring peace treaties, and
providing secure perimeters around bases.

Substraction of backgrounds, which  are captured by static cameras,
can be useful to achieve low-bit rate video compression for
transmission of rich multimedia content. The subtracted background
is transmitted once, followed by the segmented objects which are
detected.

A common element in surveillance systems is a module that performs
background subtraction to distinguish between background pixels,
which should be ignored, and foreground pixels, which should be
processed for identification or tracking. The difficulty in
background subtraction is not to differentiate, but to maintain the
background model, its representation and its associated statistics.
In particular, capturing the background in frames where the
background can change over time. These changes can be moving trees,
leaves, water flowing, sprinklers, fountains,  video screens
(billboards) just  to name a few typical examples. Other forms of
changes are weather changes like rain and snow, illumination changes
like turning on and off the light in a room and changes in daylight.
We refer to this background type as \emph{dynamic background} (DBG)
while a background without changes or with slight changes is
referred to as \emph{static background} (SBG).

In this paper, we present a new method for capturing the background.
It is based on the application of  the \emph{diffusion bases} (DB)
algorithm. Moreover, we develop real time iterative method for
background subtraction in order to separate between background and
foreground pixels while overcoming the presence of changes in the
background. The main steps of the algorithm are:

\begin{itemize}
\item Extract the background frame by dimensionality reduction via
the application of the DB algorithm.
\item Subtract the background from the input sequence.
\item Threshold the subtracted sequence.
\item Detect the foreground objects by applying %for example
\emph{depth first search} (DFS).
\end{itemize}

We propose two versions of the algorithm - one for static background
and the other for dynamic background. To handle dynamic background,
a learning process is applied to data that contains only the
background objects in order to generate a frame that extracts the
DBG. The proposed algorithm outperform current state-of-the-art
algorithms.

The rest of this paper is organized as follows: In section
\ref{sub:relatedWork}, related algorithms for background subtraction
are presented. In section \ref{sub:DimReduction},  we present the
the \emph{diffusion bases} (DB) algorithm. The main algorithm, that
is called the \emph{background substraction algorithm using
diffusion bases} (BSDB), is presented in section \ref{BSDB}. In
section \ref{EXPERIMENTAL}, we present experimental results, a
performance analysis of the BSDB algorithm and we compare it to
other background subtraction algorithms.

\section{Related work \label{sub:relatedWork}}

Background subtraction is a widely used approach for detection of
moving objects in video sequences that are captured by static
cameras. This approach detects moving objects by differentiating
between the current frame and a reference frame, often called the
background frame, or background model. In order to extract the
objects of interest, a threshold can be applied on the subtracted
frame. The background frame should faithfully represent the scene.
It should not contain moving objects. In addition,  it must be
regularly updated in order to adapt to varying conditions such as
illumination and geometry changes. This section provides a review of
the current state-of-the-art background subtraction techniques.
These techniques range from simple approaches, aiming to maximize
speed and minimizing the memory requirements, to more sophisticated
approaches, aiming to achieve the highest possible accuracy under
any possible circumstances. The goal of these approaches is to run
in real-time. Additional references can be found in
\cite{CLK00,P04,M00}.

\begin{description}

\item {\textbf{Temporal median filter} - }

In \cite{LV01}, is was proposed to use the median value of the last
$n$ frames as the background model. This provides an adequate
background model even if the $n$ frames are subsampled with respect
to the original frame rate by a factor of 10 \cite{CGPP03}. The
median filter is computed on a special set of values that contains
the last $n$ subsampled frames and the last computed median value.
This combination increases the stability of the background model
\cite{CGPP03}.

%The main disadvantage of a median-based approach is that its
%computation requires a buffer with the recent pixel values.
A fundamental shortcoming of the the median-based approach is
the need to store the recent pixel values in order to facilitate the median computation.
Moreover, the median filter can not be described by rigorous
statistics and does not provide a deviation measure with which the
subtraction threshold can be adapted.

\item {\textbf{Gaussian average} - }

This  approach models the background independently at each pixel
location $(i,j)$ \cite{WADP97}. The model is based on ideally
fitting a Gaussian probability density function (pdf) to the last
$n$ pixels. At each new frame at time $t$, a running average is
computed by $ \psi_{t}=\alpha I_{t}+(1-\alpha)\psi_{t-1}$  where
$I_{t}$ is the current frame, $\psi_{t-1}$ is the previous average
and $\alpha$ is an empirical weight that is often chosen as a
tradeoff between stability and quick update.

In addition to speed, the advantage of the running average is given
by a low memory requirement. Instead of a buffer with the last $n$
pixel values, each pixel is classified using two parameters
$(\psi_{t},\sigma_{t})$, where $\sigma_{t}$ is the standard
deviation. Let $p_{i,j}^{t}$ be the $(i,j)$ pixel at time $t$.
$p_{i,j}^{t}$ is classified as a foreground pixel if
$|p_{i,j}^{t}-\psi_{t-1}|>k\sigma_{t}$. Otherwise $p_{i,j}^{t}$ is
classified as background pixel.

\item {\textbf{Mixture of Gaussians} - }

In order to cope with rapid changes in the background, a
multi-valued background mode was suggested in \cite{SG99}. In this
model, the probability of observing a certain pixel $x$ at time $t$
is represented by a mixture of $k$ Gaussians distributions:
$P(x_{t})=\Sigma_{i=1}^{k}w_{i,t}\eta(x_{t},\psi_{i,t},\Sigma_{i,t})$
where for each $i$-th Gaussian in the mixture at time $t$, $w$
estimates what portion of the data is accounted for by this
Gaussian, $\psi$ is the mean value, $\Sigma$ is the covariance
matrix and $\eta$ is a Gaussian probability density function. In
practice, $k$ is set to be between 3 and 5. 
% Figure \ref{fig:GusMix} illustrated this model.

Each of the $k$ Gaussian distributions describes only one of the
observable background or foreground objects. The distributions are
ranked according to the ratio between their peak amplitude $w_{i}$
and their standard deviation $\sigma_{i}$. Let $Th$ be the threshold
value. The first $B$ distributions that satisfy
$\Sigma_{i=1}^{B}w_{i}>Th$ are accepted as background. All the other
distributions are considered as foreground.

Let $I_{t}$ be a frame at time $t$. At each frame $I_{t}$, two
events take place simultaneously: assigning the new observed value
$x_{t}$ to the best matching distribution and estimating the updated
model parameters. The distributions are ranked and the first that
satisfies $(x_{t}-\psi_{i,t})/\sigma_{i,t}>2.5$ is a match for
$x_{t}$.

%\begin{figure}
%\centering
%\includegraphics[width=8cm, height=5cm]{2_2.jpg}
%\caption{Top: Visualization of the Gaussian normal models at pixel
%levels. Bottom: Three input frames at time
%$t_{0}$,$t_{1}$ and $t_{2}$ (taken from
%\cite{SG99}).}%
%\label{fig:GusMix}
%\end{figure}

\item {\textbf{Kernel density estimation (KDE)} - }

This approach models the background distribution by a non-parametric
model that is based on a Kernel Density Estimation (KDE) of the
buffer of the last \emph{n} background values (\cite{EHD00}). KDE
guarantees a smooth, continuous version of the histogram of the most
recent values that are classified as background values. This
histogram is used to approximate the background pdf.

The background pdf is given as a sum of Gaussian kernels centered at
the most recent $n$ background values, $x_{t}$:
$P(x_{t})=\frac{1}{n}\Sigma_{i=1}^{n}\eta(x_{t}-x_{i},\Sigma_{t})$
where $\eta$ is the kernel estimator function and $\Sigma_{t}$
represents the kernel function bandwidth. $\Sigma$ is estimated by
computing the median absolute deviation over the sample for
consecutive intensity values of the pixel. Each Gaussian describes
just one sample data. The buffer of the background values is
selectively updated in a FIFO order for each new frame $I_{t}$.

In this application two similar models are concurrently used, one
for long-term memory and the other for short-term memory. The
long-term model is updated using a \emph{blind} update mechanism
that prevents incorrect classification of background pixels.

\item {\textbf{Sequential kernel density approximation} - }

%Mean-shift vector techniques have been employed for various pattern
%recognition problems such as image segmentation and tracking
%(\cite{C03,CM02}). The mean-shift vector is an effective technique
%capable of directly detecting the main modes of the pdf from the
%sample data using a minimum set of assumptions.
Mean-shift vector techniques have been proved to be an effective tool for solving a variety of pattern recognition problems e.g. tracking and segmentation (\cite{C03,CM02}). One of the main advantages of these techniques is their ability to directly detect the main modes of the pdf from the sample data while relying on a minimal set of assumptions.
Unfortunately, the computational cost of this approach is very high.
%Furthermore, due to the iterative nature of this method, its convergence needs to be investigated.
%However, it has a
%very high computational cost since it is an iterative technique and
%it requires a study of the convergence over the whole data space.
As such, it is not immediately applicable to modeling background pdfs
at the pixel level.

To solve this problem, computational optimizations are used to
mitigate the computational high cost (\cite{PJ04}). Moreover, the
mean-shift vector can be used only for an off-line model
initialization \cite{JCD04}, i.e. the initial set of Gaussian modes
of the background pdf is detected from an initial sample set. The
real-time model is updated by simple heuristics that handle
 mode adaptation, creations, and merging.

\item {\textbf{Co-occurrence of image variations} - }

This method exploits spatial cooccurrences of image variations
(\cite{SWFS03}). It assumes that neighboring blocks of pixels that
belong to the background should have similar variations over time.
The disadvantage of this method is that it does not handle blocks at
the borders of distinct background objects.

This method divides each frame to distinct blocks of $N\times N$
pixels where each block is regarded as an $N^{2}$-component vector.
This trades-off resolution with high speed and better stability.
During the learning phase, a certain number of samples is acquired
at a set of points, for each block. The temporal average is computed
and the differences between the samples and the average, called the
\emph{image variations}, is calculated. Then the $N^{2}\times N^{2}$
covariance matrix is computed with respect to the average. An
eigenvector transformation is applied to reduce the dimensions of
the image variations.

For each block $b$, a classification phase is performed: the
corresponding current eigen-\emph{image-variations} are computed on
a neighboring block of $b$. Then the image variation is expressed as
a linear interpolation of its L-nearest neighbors in the eigenspace.
The same interpolation coefficients are applied on the values of
$b$, to provide an estimate for its current
eigen-\emph{image-variations}.

\item {\textbf{Eigen-backgrounds}  - }

This approach is based on an eigen-decomposition of the whole image
\cite{ORP00}. During a learning phase, samples of $n$ images are
acquired. The average image is then computed and subtracted from all
the images. The covariance matrix is computed and the best
eigenvectors are stored in an eigenvector matrix. For each frame
$I$, a classification phase is executed: $I$ is projected onto the
eigenspace and then projected back onto the image space. The output
is the background frame, which does not contain any small moving
objects. A threshold is applied on the difference between $I$ and
the background frame. 
%An experimental result is shown in Fig. \ref{fig:EigenBg}.

%\begin{figure}
%\begin{center}
%\includegraphics[width=14cm]{2_3.jpg}
%\end{center}
%\caption{Left: The input frame. Middle: The background frame. Right:
%The output frame. (taken from \cite{ORP00})}%
%\label{fig:EigenBg}
%\end{figure}

\end{description}

\section{Dimensionality reduction \label{sub:DimReduction}}

Dimensionality reduction has been extensively researched. Classic
techniques for dimensionality reduction such as Principal Component
Analysis (PCA) and Multidimensional Scaling (MDS) are simple to
implement and can be efficiently computed. However, they guarantee
to discover the true structure of a data set only when the data set
lies on or near a linear subspace of the high-dimensional input
space (\cite{MKB02}). These methods are highly sensitive to noise
and outliers since they take into account the distances between
\emph{all} pairs of points. Furthermore, PCA and MDS fail to detect
non-linear structures.

More recent dimensionality reduction methods like Local Linear
Embedding (LLE) \cite{LLE00} and ISOMAP \cite{ISO00} amend this
pitfall by considering for each point only the distances to its
closest neighboring points in the data. Recently, Coifman and Lafon
\cite{CL_DM04} introduced the \emph{Diffusion Maps} (DM) algorithm
which is a manifold learning scheme. DM embeds high dimensional data
into an Euclidean space of substantially smaller dimension while
preserving the geometry of the data set. The global geometry is
preserved by maintaining the local neighborhood geometry of each
point in the data set. DM uses a random walk distance that is more
robust to noise since it averages all the paths between a pair of
points.
%Furthermore, DM can provide parameterization of the data
%when only a point-wise similarity matrix is available. This may
%occur either when the original data consists of abstract objects or
%when there is no access to the original data.

\emph{Diffusion Bases} (DB) - a dual algorithm to the DM algorithm -
is described in (A. Schclar and A. Averbuch. "Segmentation and anomalies detection in hyper-spectral images via diffusion bases", preprint, 2008). The DB algorithm is dual to the DM
algorithm in the sense that it explores the variability among the
\emph{coordinates} of the original data. Both algorithms share a
graph Laplacian construction, however, the DB algorithm uses the
Laplacian eigenvectors as an orthonormal system on which it
projects the original data.% on it. %The DB algorithm also assumes that there is

\subsection{Diffusion Bases (DB) \label{sub:db}}

This section reviews the DB algorithm for dimensionality reduction.
Let $\Omega=\left\{ x_{i}\right\} _{i=1}^{m},\,
x_{i}\in\mathbb{R}^{n}$, be a data set and let $x_{i}\left(j\right)$
denote the $j^{th}$ coordinate of $x_{i}$, $1\le j\le n$. We define
the vector
$y_{j}\triangleq\left(x_{1}\left(j\right),\dots,x_{m}\left(j\right)\right)$
as the vector whose components are composed of the $j^{th}$
coordinate of all the points in $\Omega$.
%as it is described in \cite{SA07}:
The DB algorithm consists of the following steps:

\begin{itemize}

\item Construct the data set $\Omega'=\left\{ y_{j}\right\}
_{j=1}^{n}$

\item Build a non-directed graph $G$ whose vertices correspond to $\Omega'$ with a
non-negative and fast-decaying weight function $w_{\varepsilon}$
that corresponds to the \emph{local} point-wise similarity between
the points in $\Omega'$. By fast decay we mean that given a scale
parameter $\varepsilon>0$ we have
$w_{\varepsilon}\left(y_{i},y_{j}\right)\rightarrow0$ when
$\left\Vert y_{i}-y_{j}\right\Vert \gg\varepsilon$ and
$w_{\varepsilon}\left(y_{i},y_{j}\right)\rightarrow1$ when
$\left\Vert y_{i}-y_{j}\right\Vert \ll\varepsilon$. %A proper choice
%of $\varepsilon$ produces a sparse $w_{\varepsilon}$.
%For every pair $x_{i},x_{j}\in\Omega$, the
%weight function has the following properties. Symmetry:
%$w_{\varepsilon}\left(x_{i},x_{j}\right)=w_{\varepsilon}\left(x_{j},x_{i}\right)$.
%Non-negativity: $w_{\varepsilon}\left(x_{i},x_{j}\right)\ge0$. Fast
%decay:
One of the common choices for $w_{\varepsilon}$ is
\begin{equation}
w_{\varepsilon}\left(y_{i},y_{j}\right)=\exp\left(-\frac{\left\Vert
y_{i}-y_{j}\right\Vert ^{2}}{\varepsilon}\right)\label{eq:gaussian}
\end{equation}
where $\varepsilon$ defines a notion of neighborhood by defining a
$\varepsilon$-neighborhood for every point $y_{i}$.

\item Construction of a random walk on the graph $G$ via a Markov transition
matrix $P$. $P$ is the row-stochastic version of $w_{\varepsilon}$
which is derived by dividing each row of $w_{\varepsilon}$ by its
sum\footnote{$P$ and the graph Laplacian $I-P$ (see \cite{C97})
share the same eigenvectors.}.

\item Perform an eigen-decomposition of $P$ to produce the left and the right
eigenvectors of $P$: $\left\{ \psi_{k}\right\} _{k=1,\dots,n}$ and
$\left\{ \xi_{k}\right\} _{k=1,\dots,n}$, respectively. Let $\left\{
\lambda_{k}\right\} _{k=1,\dots,n}$ be the eigenvalues of $P$ where
$\left|\lambda_{1}\right|\ge\left|\lambda_{2}\right|\ge...\ge\left|\lambda_{n}\right|$.

%\item The DM algorithm is applied on the set of vectors
%$\Omega'=\left\{y_{j}\right\} j=1,...,n$.

\item The right eigenvectors
of $P$ constitute an orthonormal basis $\left\{ \xi_{k}\right\}
_{k=1,\dots,n},\,\xi_{k}\in\mathbb{R}^{n}$. These eigenvectors
capture the \emph{non-linear} coordinate-wise variability of the
original data.

\item Next, we use the spectral decay property of the spectral decomposition
to extract only the first $\eta$ eigenvectors %(we are not excluding
%the first eigenvector as mentioned in section
%\ref{sub:Spectral-decomposition}),
$BS\triangleq\left\{
\xi_{k}\right\} _{k=1,\dots,\eta},$ which contain the
\emph{non-linear} directions with the highest variability of the
coordinates of the original data set $\Omega$.
\item We project the original data $\Omega$ onto the basis $BS$. Let $\Omega_{BS}$
be the set of these projections: $\Omega_{BS}=\left\{ g_{i}\right\}
_{i=1}^{m},\, g_{i}\in\mathbb{R}^{\eta},$ where
$g_{i}=\left(x_{i}\cdot\xi_{1},\dots,x_{i}\cdot\xi_{\eta}\right),\,\,
i=1,\dots,m$ and $\cdot$ denotes the inner product operator.
$\Omega_{BS}$ contains the coordinates of the original points in the
orthonormal system whose axes are given by $BS$. Alternatively,
$\Omega_{BS}$ can be interpreted in the following way: the
coordinates of $g_{i}$ contain the correlation between $x_{i}$ and
the directions given by the vectors in $BS$.
\end{itemize}
A summary of the DB procedure is given in Algorithm
\ref{alg:Diffusion-Basis-Calculation}. An enhancement of the
spectral decomposition is described in (A. Schclar and A. Averbuch. "Segmentation and anomalies detection in hyper-spectral images via diffusion bases", preprint, 2008).

\begin{algorithm}
\textbf{DiffusionBasis($\Omega'$, $w_{\varepsilon}$, $\varepsilon$,
$\eta$)}

\begin{enumerate}
\item Calculate the weight function $w_{\varepsilon}\left(y_{i},y_{j}\right),\,\, i,j=1,\dots n$,
(Eq. \ref{eq:gaussian}).
\item Construct a Markov transition matrix $P$ by normalizing the sum of
each row in $w_{\varepsilon}$ to be 1: \[
p\left(y_{i},y_{j}\right)=\frac{w_{\varepsilon}\left(y_{i},y_{j}\right)}{d\left(y_{i}\right)}\]
where
$d\left(y_{i}\right)=\sum_{j=1}^{n}w_{\varepsilon}\left(y_{i},y_{j}\right)$.
\item Perform a spectral decomposition of $P$\[
p\left(y_{i},y_{j}\right)\equiv\sum_{k=1}^{n}\lambda_{k}\xi_{k}\left(y_{i}\right)\psi_{k}\left(y_{j}\right)\]
where the left and the right eigenvectors of $P$ are given by
$\left\{ \psi_{k}\right\} $ and $\left\{ \xi_{k}\right\} $,
respectively, and $\left\{ \lambda_{k}\right\} $ are the eigenvalues
of $P$ in descending order of magnitude.
\item Let $BS\triangleq\left\{ \xi_{k}\right\} _{k=1,\dots,\eta}.$
\item Project the original data $\Omega$ onto the orthonormal system $BS$:\[
\Omega_{BS}=\left\{ g_{i}\right\} _{i=1}^{m},\,
g_{i}\in\mathbb{R}^{\eta}\] where\[
g_{i}=\left(x_{i}\cdot\xi_{1},\dots,x_{i}\cdot\xi_{\eta}\right),\,\,
i=1,\dots,m,\,\xi_{k}\in BS,\,1\le k\le\eta\]
 and $\cdot$ is the inner product.
\item \textbf{return} $\Omega_{BS}$.
\end{enumerate}

\caption{The Diffusion Basis algorithm. \label{alg:Diffusion-Basis-Calculation}}
\end{algorithm}

\section{The Background Subtraction Algorithm using Diffusion Bases
(BSDB)} \label{BSDB}

In this section we present the BSDB algorithm. The algorithm has two
versions:
\begin{description}
\item {\textbf{Static background subtraction using DB (SBSDB)}:}
 We assume that the background is static
(SBG) -- see section \ref{SBG}. The video sequence is captured
on-line. Gray level images are sufficient for the processing.
\item {\textbf{Dynamic background substraction using DB (DBSDB)}:}
 We assume
that the background is moving (DBG) -- see section \ref{DBG}. This
algorithm uses off-line (training) and on-line (detection)
procedures. As opposed to the SBSDB, this algorithm requires color
 (RGB) frames.
\end{description}

We assume that in both algorithms the camera is static.

\subsection{Static background subtraction algorithm using DB (SBSDB)}
\label{SBG}

In this section we describe the on-line algorithm that is applied on
a video sequence that is captured by a static camera. We assume that
the background is static. The SBSDB algorithm captures the static
background, subtracts it from the video sequence and segments the
subtracted output.

The input to the algorithm is a sequence of video frames in
gray-level format. The algorithm produces a binary mask for each
video frame. The pixels in the binary mask that belong to the
background are assigned 0 values while the other pixels are assigned
to be 1.

%\paragraph*{Off-line algorithm for capturing static background\label{sub:findSteadyBg}}
\subsubsection{Off-line algorithm for capturing static background\label{sub:findSteadyBg}}

In order to capture the static background of a scene, we reduce the
dimensionality of the input sequence by applying the DB algorithm
(Algorithm \ref{alg:Diffusion-Basis-Calculation} in section
\ref{sub:db}). The input to the algorithm consists of $n$ frames
that form a datacube.

Formally, let $D_{n}=\left\{s_{i,j}^{t}, i,j=1,...,N,
t=1,...,n\right\}$ be the input datacube of $n$ frames each of size
$N\times N$ where $s_{i,j}^{t}$ is the pixel at position $(i,j)$ in
the video frame at time $t$. We define the vector
$P_{i,j}\triangleq\left(s_{i,j}^{1},...,s_{i,j}^{n}\right)$ to be
the values of the $(i,j)^{th}$ coordinate at all the $n$ frames in
$D_{n}$. This vector is referred to as a \emph{hyperpixel}. Let
$\Omega_{n}=\left\{P_{i,j}\right\}$, $i,j=1,...,N$ be the set of all
hyperpixels. We define
$F_{t}\triangleq(s_{1,1}^{t},...,s_{N,N}^{t})$ to be a 1-D vector
representing the video frame at time $t$. We refer to $F_{t}$ as a
frame-vector. Let
$\Omega'_{n}\triangleq\left\{F_{t}\right\}_{t=1}^{n}$ be the set of
all frame-vectors.

We apply the DB algorithm to $\Omega_{n}$ in
$\Omega_{BS}$=DiffusionBasis($\Omega'_{n}$, $w_{\varepsilon}$,
$\varepsilon$, $\eta$) where $w_{\varepsilon}$ is defined by Eq.
\ref{eq:gaussian}, $\varepsilon$ and $\eta$ are defined in section
\ref{sub:db} - see Algorithm \ref{alg:Diffusion-Basis-Calculation}.
The output is the projection of every hyperpixel on the diffusion
basis which embeds the original data $D_{n}$ into a reduced space.
The first vector of $\Omega_{BS}$ represents the background of the
input frames. Let $bg_{V}=\left(x_{i}\right)$, $i=1,...,N^{2}$ be
this vector. We reshape $bg_{V}$ into the matrix
$bg_{M}=\left(x_{i,j}\right), i,j=1,...,N$. Then, $bg_{M}$ is
normalized to be between 0 to 255. The normalized background is
denoted by $\widehat{bg}_{M}$.

%\paragraph*{On-line algorithm for capturing a static background\label{sub:findBgRT} }
\subsubsection{On-line algorithm for capturing a static background\label{sub:findBgRT} }

In order to make the algorithm suitable for on-line applications,
the incoming video sequence is processed by using a \emph{sliding
window} (SW) of size $m$. Thus, the number of frames that are input
to the algorithm is $m$. Naturally, we seek to minimize $m$ in order
to obtain a faster result from the algorithm. We found empirically
that the algorithm produces good results for values of $m$ as low as
$m=5,6$ and $7$. The delay of 5 to 7 frames is negligible and
renders the algorithm to be suitable for on-line applications.

\begin{figure}
\begin{center}
\includegraphics[width=13cm]{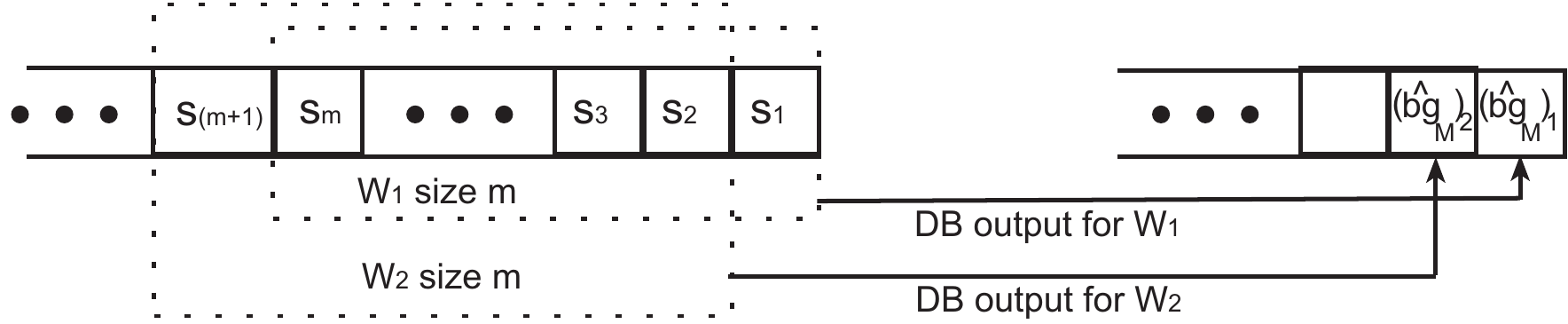}
\end{center}
\caption{Illustration of how the SW is shifted.
$W_{1}=\left(s_{1},...,s_{m}\right)$ is the SW for $s_{1}$.
$W_{2}=\left(s_{2},...,s_{m+1}\right)$ is the SW for $s_{2}$, etc.
The backgrounds of $s_{i}$ and $s_{i+1}$ are denoted by
$(\widehat{bg}_{M})_{i}$ and $(\widehat{bg}_{M})_{i+1}$, $i=1,...,n-m+1$, respectively.}%
\label{fig:slidingWindow}%
\end{figure}

Let $S=\left(s_{1},...,s_{i}...,s_{m},s_{m+1},...,s_{n}\right)$ be
the input video sequence. we apply the algorithm that is described
in section \ref{sub:findSteadyBg} to every SW. The output is a
sequence of background frames
\begin{equation}
\widehat{BG}=\left((\widehat{bg}_{M})_{1},...,(\widehat{bg}_{M})_{i},...,(\widehat{bg}_{M})_{m},(\widehat{bg}_{M})_{m+1},...,(\widehat{bg}_{M})_{n}\right)
\end{equation}
where $(\widehat{bg}_{M})_{i}$ is the background that corresponds to
frame $s_{i}$ and $(\widehat{bg}_{M})_{n-m+2}$ till
$(\widehat{bg}_{M})_{n}$ are equal to $(\widehat{bg}_{M})_{n-m+1}$.
Figure \ref{fig:slidingWindow} describes how the SW is shifted.

The SW results in a faster execution time of the DB algorithm. The
weight function $w_{\varepsilon}$ (Eq. \ref{eq:gaussian}) is not
recalculated for all the frame in the SW. Instead, $w_{\varepsilon}$
is only updated according to the new frame that enters the SW and
the one that exits the SW. Specifically, let
$W_{t}=\left(s_{t},...,s_{t+m-1}\right)$ be the SW at time $t$ and
let $W_{t+1}=\left(s_{t+1},...,s_{t+m}\right)$ be the SW at time
$t+1$. At time $t+1$, $w_{\varepsilon}$ is calculated only for
$s_{t+m}$ and the entries that correspont to $s_{t}$ are removed
from $w_{\varepsilon}$.

%\paragraph*{The SBSDB algorithm\label{sub:algSteadyBg}}
\subsubsection{The SBSDB algorithm\label{sub:algSteadyBg}}

The SBSDB on-line algorithm captures the background of each SW
according to section \ref{sub:findBgRT}. Then it subtracts the
background from the input sequence and thresholds the output to get
the background binary mask.

Let $S=\left(s_{1}...,s_{n}\right)$ be the input sequence. For each
frame $s_{i}\in S$, $i=1,...,n$, we do the following:
\begin{itemize}
\item Let $W_{i}=\left(s_{i},...,s_{i+m-1}\right)$ be the SW of $s_{i}$. The
on-line algorithm for capturing the background (section
\ref{sub:findBgRT}) is applied to $W_{i}$. The output is the
background frame $(\widehat{bg}_{M})_{i}$.
\item The SBSDB algorithm subtracts $(\widehat{bg}_{M})_{i}$ from the original input
frame by $\bar{s}_{i}=s_{i}-(\widehat{bg}_{M})_{i}$. Then, each
pixel in $\bar{s}_{i}$ that has a negative value is set to 0.
\item A threshold is applied to $\bar{s}_{i}$. The threshold is computed
in section \ref{sub:GrayThreshold}. For $k,l=1,...,N$ the output is
defined as follows:
\[\tilde{s}_{i}(k,l)=\left\{
                                         \begin{array}{ll}
                                           0, & \hbox{if it is a background pixel;} \\
                                           1, & \hbox{otherwise.}
                                         \end{array}
                                       \right.\]
\end{itemize}

%\paragraph*{Threshold computation for a grayscale input\label{sub:GrayThreshold}}
\subsubsection{Threshold computation for a grayscale input\label{sub:GrayThreshold}}

The threshold $Th$, which separates between background and
foreground pixels, is calculated in the last step of the SBSDB
algorithm. The SBSDB algorithm subtracts the background from the
input frame and sets pixels with negative values to zero. Usually,
the histogram of a frame after subtraction will be high at small
values and low at high values. The SBSDB algorithm smooths the
histogram in order to compute the threshold value accurately.

Let $h$ be the histogram of a frame and let $\mu$ be a given
parameter which provides a threshold for the slope of $h$. $\mu$ is
chosen to be the magnitude of the slope where $h$ becomes moderate.
We scan $h$ from its global maximum to the right. We set the
threshold $Th$ to be the smallest value of $x$ that satisfies
$h'(x)<\mu$ where $h'_{x}$ is the first derivative of $h$ at point
$x$, i.e. the slope of $h$ at point $x$. The background/foreground
classification of the pixels in the input frame $\bar{s}_{i}$ is
determined according to $Th$. Specifically, for $k,l=1,...,N$
\[\tilde{s}_{i}(k,l)=\left\{
                                         \begin{array}{ll}
                                           0, & \hbox{if $\bar{s}_{i}(k,l)<Th$;} \\
                                           1, & \hbox{otherwise.}
                                         \end{array}
                                       \right.\]

\begin{figure}
\begin{center}
\includegraphics[width=5cm]{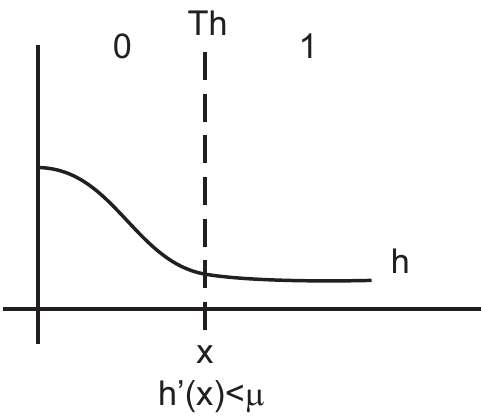}
\end{center}
\caption{An example how to  use the histogram $h$ for finding the
threshold value. $Th$ is set to $x$ since $h'(x)<\mu$.}%
\label{fig:hist_gray}
\end{figure}

Fig. \ref{fig:hist_gray} illustrates how to find the threshold.

%-----------------------------------

\subsection{Dynamic background subtraction algorithm using DB
(DBSDB)} \label{DBG}

%The DBSDB is an two-step on-line algorithm that is used for video
%sequences in which the background is dynamic (moving). The first
%step consists of an offline training procedure that captures the
%dynamic background of the scene from a video sequence that does not
%contain foreground objects. The second step captures the dynamic
%background, subtracts it from the and the background that was
%captured during the first step from applies an on-line background
%subtraction
In this section, we describe an on-line algorithm that handles video
sequences that are captured by a static camera. We assume that the
background is dynamic (moving). The DBSDB applies an off-line
procedure that captures the dynamic background and an on-line
background subtraction algorithm. In addition, the DBSDB algorithm
segments the video sequence after the background subtraction is
completed.

\begin{figure}
\begin{center}
\includegraphics[width=8cm]{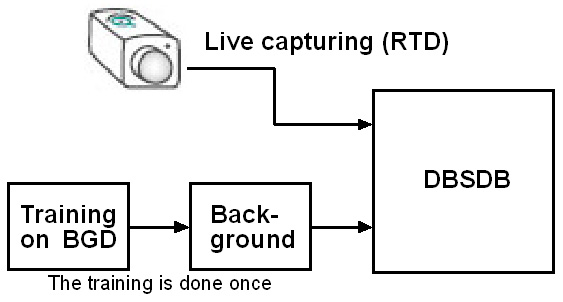}
\end{center}
\caption{The inputs to the DBSDB algorithm. The training is done
once on the BGD. It produces the background which is input to
the DBSDB. The RTD is the on-line input to the DBSDB.}%
\label{fig:input}
\end{figure}

The input to the algorithm consists of two components:
\begin{description}
\item {\textbf{Background training data}: }
A video sequence of the scene without foreground objects. This
training data can be obtained from the frames in the beginning of
the video sequence.
This sequence is referred to as the \emph{background data} (BGD). %A small
%number of frames is sufficient.
\item {\textbf{Data for classification}: }
A video sequence that contains background and foreground objects.
The classification of the objects is performed on-line. We refer to
this sequence as the \emph{real-time data} (RTD).
\end{description}
For both input components, the video frames are assumed to be in RGB
-- see Fig. \ref{fig:input}.

The algorithm is applied to every video frame and a binary mask is
constructed in which the pixels that belong to the background are
set to 0 while the foreground pixels are set to 1.

%\paragraph*{Iterative method for capturing a dynamic background: training \label{sub:captureDynamicBg}}
\subsubsection{Iterative method for capturing a dynamic background - training \label{sub:captureDynamicBg}}

The algorithm that is described in section \ref{sub:algSteadyBg},
does not handle well on-going changes in the background, such as
illumination differences between frames, moving leaves, water
flowing, etc. In the following, we present a method that is not
affected by background changes. An iterative procedure is applied on
the BGD in order to capture the movements in the scene. This
procedure constitutes the training step of the algorithm.

Let $B=\left(b_{1},...,b_{m}\right)$ be the BGD input sequence and
let $bg_{M}^{final}$ be the output background frame.
$bg_{M}^{final}$ is initialized to zeros. Each iteration contains
the following steps:
\begin{itemize}
\item Application of the off-line algorithm (section \ref{sub:findSteadyBg})
in order to capture the static background of $B$. The BGD is treated
as a single sliding window of length $m$. The output consists of the
background frames $bg_{M}$ and $\widehat{bg}_{M}$ where
$\widehat{bg}_{M}$ is the normalization of $bg_{M}$.
\item $\widehat{bg}_{M}$ is subtracted from each frame in $B$ by $\bar{b}_{j}=b_{j}-\widehat{bg}_{M}$, $j=1,...,m$.
In case the input is in grayscale format, we set to zero each pixel
in $\bar{b}_{j}$ that has a negative value. The output is the
sequence $\bar{B}=\left(\bar{b}_{1},...,\bar{b}_{m}\right)$.
\item $bg_{M}$ is added to $bg_{M}^{final}$ by $bg_{M}^{final}=bg_{M}^{final}+bg_{M}$.
\item $\bar{B}$ is the input for the next iteration, $B=\bar{B}$.
\end{itemize}

The iterative process stops when a given number of pixels in $B$ are
equal to or smaller than zero. Finally, $bg_{M}^{final}$ is
normalized to be between 0 to 255. The normalized background is
denoted by $\widehat{bg}_{M}^{final}$. The output of this process is
composed of $bg_{M}^{final}$ and $\widehat{bg}_{M}^{final}$.

%\paragraph*{The DBSDB algorithm \label{sub:algDynamicBg}}
\subsubsection{The DBSDB algorithm \label{sub:algDynamicBg}}

In this section, we describe the DBSDB algorithm which handles video
sequence that contain a dynamic background. The DBSDB algorithm
consists of a training phase, which captures the BGD (section
\ref{sub:captureDynamicBg}), and a classification phase, which is
applied on the RTD. Both phases process grayscale and RGB versions
of the input and generate grayscale and RGB outputs. The final phase
combines the output from the grayscale classification phase and the
output from the RGB classification phase.

Formally, let $S^{rgb}=\left(s_{1}^{rgb},..,s_{n}^{rgb}\right)$ and
$B^{rgb}=\left(b_{1}^{rgb},...,b_{m}^{rgb}\right)$ be the RTD, which
is the on-line captured video sequence, and the BGD, which is the
off-line video sequence for the training phase (section
\ref{sub:captureDynamicBg}), respectively.\\

The DBSDB algorithm consists of the following:

\begin{enumerate}
%\begin{itemize}
\item {\textbf{The grayscale training phase}}
\par\noindent
\begin{itemize}
\item  Convert $B^{rgb}$ into grayscale format. The grayscale sequence
is denoted by $B^{g}$.
\item Apply the SBSDB algorithm to $B^{g}$ \emph{excluding} the threshold
computation, as was done in section \ref{sub:algSteadyBg}. The
output is a sequence of background frames $\bar{B}^{g}$.
\item Capture the \emph{dynamic background} (DBG) in $\bar{B}^{g}$ (section \ref{sub:captureDynamicBg}).
The output is the background frame given by
$\widehat{(bg}_{M}^{final})^{g}$.
\end{itemize}

\item {\textbf{The RGB training phase:}}
\par\noindent
Capture the DBG in each of the RGB channels of $B^{rgb}$ (section
\ref{sub:captureDynamicBg}). The output is the background frame
denoted by $\widehat{(bg}_{M}^{final})^{rgb}$.

\item {\textbf{The grayscale classification phase: }}
\par\noindent
 $S^{rgb}$ is converted into grayscale format. The grayscale
sequence is denoted by $S^{g}$. The SBSDB algorithm is applied on
$S^{g}$ \emph{excluding} the threshold computation, as it is
described in section \ref{sub:algSteadyBg}. This process is
performed once or, in some cases, iteratively twice. The output is
denoted by $\bar{S}^{g}$.

For each frame $\bar{s}_{i}^{g}\in \bar{S}^{g}$, $i=1,...,n$, we do
the following:
\begin{itemize}
\item  $\widehat{(bg}_{M}^{final})^{g}$ is subtracted from $\bar{s}_{i}^{g}$ by $\tilde{s}_{i}^{g}=\bar{s}_{i}^{g}-\widehat{(bg}_{M}^{final})^{g}$.
Then, each pixel in $\tilde{s}_{i}^{g}$ that has a negative value is
set to 0.
\item A threshold is applied to $\tilde{s}_{i}^{g}$. The threshold is computed
as in section \ref{sub:GrayThreshold}. The output is set to:
\[\breve{s}_{i}^{g}(k,l)=\left\{
                                         \begin{array}{ll}
                                           0, & \hbox{if it is a background pixel;} \\
                                           1, & \hbox{otherwise.}
                                         \end{array}
                                       \right.\]
for $k,l=1,...,N$.
\end{itemize}

\item {\textbf{The RGB classification phase:}}
\par\noindent
 For each frame $s_{i}^{rgb}\in S^{rgb}$, $i=1,...,n$, we do the
following:
\begin{itemize}
\item  $\widehat{(bg}_{M}^{final})^{rgb}$ is subtracted from $s_{i}^{rgb}$ by $\bar{s}_{i}^{rgb}=s_{i}^{rgb}-\widehat{(bg}_{M}^{final})^{rgb}$.
\item $\bar{s}_{i}^{rgb}$ is normalized to be between 0 to 255. The normalized frame is denoted by $\tilde{s}_{i}^{rgb}$.
\item A threshold is applied to $\tilde{s}_{i}^{rgb}$. The threshold is computed according to
section \ref{sub:RGBThreshold}. The output is set to:
\[\breve{s}_{i}^{rgb}(k,l)=\left\{
                                         \begin{array}{ll}
                                           0, & \hbox{if it is a background pixel;} \\
                                           1, & \hbox{otherwise.}
                                         \end{array}
                                       \right.\]
for $k,l=1,...,N$.
\end{itemize}

\item {\textbf{The DFS phase:}}
\par\noindent
This phase combines the $\breve{s}_{i}^{g}$ and
$\breve{s}_{i}^{rgb}$ from the grayscale and RGB classification
phases, respectively.  Since $\breve{s}_{i}^{g}$ contains false
negative detections (not all the foreground objects are found) and
$\breve{s}_{i}^{rgb}$ contains false positive detections (background
pixels are classified as foreground pixels), we use each foreground
pixel in $\breve{s}_{i}^{g}$ as a reference point from which we
begin the application of a DFS on $\breve{s}_{i}^{rgb}$ (see section
\ref{sub:dfs}).
\end{enumerate}
%\end{itemize}

\paragraph*{Threshold computation for RGB input\label{sub:RGBThreshold}}

In the last step of the RGB classification phase in the DBSDB
algorithm, the thresholds that separate between background pixels
and foreground pixels are computed for each of the RGB components.
The DBSDB algorithm subtracts the background from the input frame,
therefore, the histogram of a frame after the subtraction is high in
the center and low at the right and left ends, where the center area
corresponds to the background pixels. The DBSDB algorithm smooths
the histogram in order to compute the threshold values accurately.

Let $h$ be the histogram and let $\mu$ be a given parameter which
provides a threshold for the slope of $h$. $\mu$ should be chosen to
be the value of the slope where $h$ becomes moderate. We denote the
thresholds to be $Th^{r}$ and $Th^{l}$. We scan $h$ from its global
maximum to the right. $Th^{r}=x$ if $x$ is the first coordinate that
satisfies $h'(x)<\mu$ where $h'(x)$ denotes the first derivative of
$h$ at point $x$, i.e. the slope of $h$ at point $x$. We also scan
$h$ from its global maximum to the left. $Th^{l}=y$ if $y$ is the
first coordinate that satisfies $h'(y)>-\mu$.

\begin{figure}[H]
\begin{center}
\includegraphics[width=7cm]{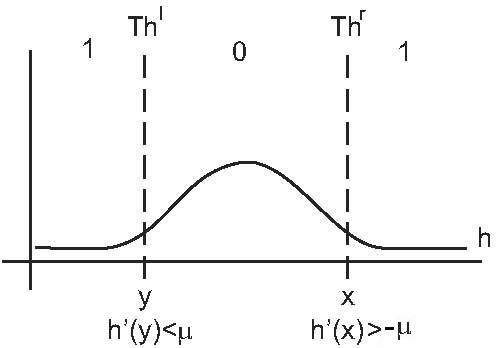}
\end{center}
\caption{An example that uses the histogram $h$ for finding the
threshold values. %$Th^{r}$
%and $Th^{l}$ are set to $x$ and $y$,
%respectively, where the slope of $h$ becomes smaller than $\mu$.
}
\label{fig:hist_rgb}%
\end{figure}

The classification of the pixels in the input frame
$\tilde{s}_{i}^{rgb}$ is determined according to $Th^{r}$ and
$Th^{l}$. For each color component and for each $k,l=1,...,N$
\[\breve{s}_{i}^{rgb}(k,l)=\left\{
                                         \begin{array}{ll}
                                           0, & \hbox{if $Th^{l}<\tilde{s}_{i}^{rgb}(k,l)<Th^{r}$;} \\
                                           1, & \hbox{otherwise.}
                                         \end{array}
                                       \right.\]
See Fig.\ref{fig:hist_rgb} for an example how the thresholds are
computed.

The process is executed three times, one for each of the RGB
channels. The outputs are combined by a pixel-wise OR operation.

\paragraph*{Scan by depth-first search (DFS) \label{sub:dfs}}

The last phase of the DBSDB algorithm is the application of a DFS.
Let $\breve{s}_{i}^{g}\in \breve{S}^{g}$ and $\breve{s}_{i}^{rgb}\in
\breve{S}^{rgb}$ be the $i^{th}$ output frames of the grayscale and
the RGB classification phases, respectively. Each frame is a binary
mask represented by a matrix. The DFS phase combines both outputs.
In $\breve{s}_{i}^{g}$ there are false negative detections and in
$\breve{s}_{i}^{rgb}$ there are false positive detections. We use
each foreground pixel in $\breve{s}_{i}^{g}$ as a reference point
from which we begin a DFS in $\breve{s}_{i}^{rgb}$. The goal is to
find the connected components of the graph whose vertices are
constructed from the pixels in $\breve{s}_{i}^{rgb}$ and whose edges
are constructed according to the 8-neighborhood of each pixel.

\begin{figure}[H]
\begin{center}
\includegraphics[width=9cm]{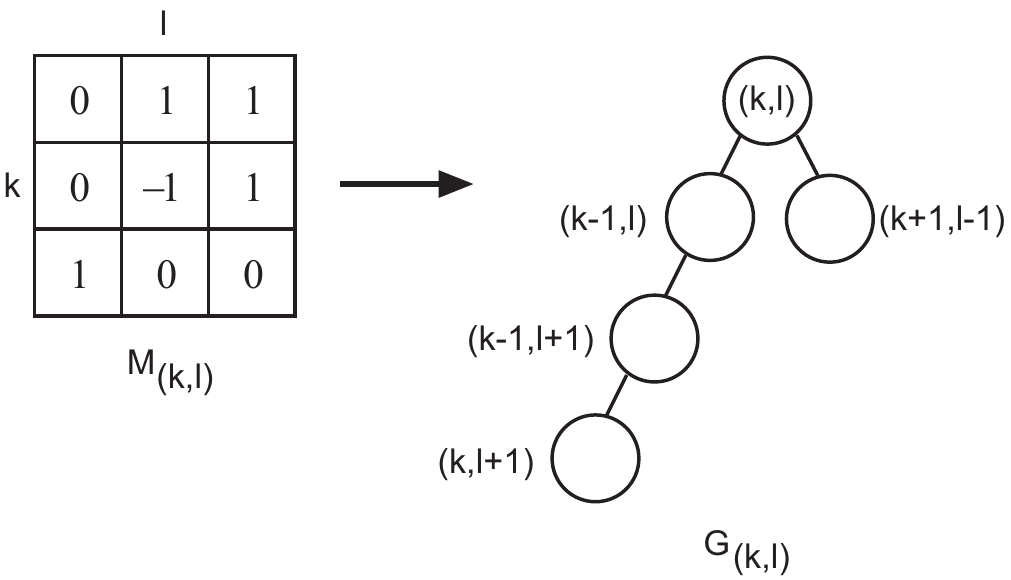}
\end{center}
\caption{$G_{(k,l)}$ is a graph representation of the neighboring
matrix $M_{(k,l)}$ of the root pixel $\breve{s}_{i}^{rgb}(k,l)$. A
root pixel is set to -1, a foreground pixel is set to 1 and a
background pixel is set to 0.} \label{fig:dfs}
\end{figure}

The graph is constructed as follows:
\begin{itemize}
\item A pixel $\breve{s}_{i}^{rgb}(k,l)$ is a root if
$\breve{s}_{i}^{g}(k,l)$ is a foreground pixel and it has not been
classified yet as a foreground pixel by the algorithm.
\item A pixel $\breve{s}_{i}^{rgb}(k,l)$ is a node if
it is a foreground pixel and was not marked yet as a root.
\item Let $\breve{s}_{i}^{rgb}(k,l)$ be a node or a root and let
$M_{(k,l)}$ be a $3\times 3$ matrix that represent its
8-neighborhood. A pixel $\breve{s}_{i}^{rgb}(q,r)\in M_{(k,l)}$ is a
child of $\breve{s}_{i}^{rgb}(k,l)$ if $\breve{s}_{i}^{rgb}(q,r)$ is
a node (see Fig.\ref{fig:dfs}).
\end{itemize}

The DFS is applied from each root in the graph. Each node, that is
scanned by the DFS, represents a pixel that belongs to the
foreground objects that we wish to find. The scanned pixels are
marked as the new foreground pixels and the others as the new
background pixels.

\subsection{A parallel extension of the SBSDB and the DBSDB algorithms}

We propose parallel extensions to the SBSDB and the DBSDB
algorithms. We describe this scheme for the SBSDB algorithm and the
same scheme can be used for the DBSDB algorithm.

First, the data cube $D_{n}=\left\{
s_{i,j}^{t},i,j=1,...,N,t=1,...,n\right\} $ is decomposed into
overlapping blocks $\left\{ \beta_{k,l}\right\} $. Next, the SBSDB
algorithm is independently applied on each block. This step can run
in \emph{parallel}. The final result of the algorithm is constructed
using the results from each block. Specifically, the result from
each block is placed at its original location in $D_{n}$. The result
for pixels that lie in overlapping areas between adjacent blocks is
obtaind by applying a logical $OR$ operation on the corresponding
blocks results.

%----------------------------

\section{Experimental results } \label{EXPERIMENTAL}

In this section, we present the results from the application of the
SBSDB and DBSDB algorithms. The section is divided into three parts:
%(a) A comparison between the outputs obtained by the application of
%the DB based algorithm and the outputs that are obtained when the DB
%algorithm is replaced by \emph{principal components analysis} (PCA).
The first part is composed from the results of the SBSDB algorithm
when applied to a SBG video. The second part contains the results
from the application of the DBSDB algorithm to a DBG video. In the
third part we compare between the results obtained by our algorithm
and those obtained by five other background-subtraction algorithms.

\subsection{Performance analysis of the SBSDB algorithm}

\begin{figure}
\begin{center}
\includegraphics[width=13cm]{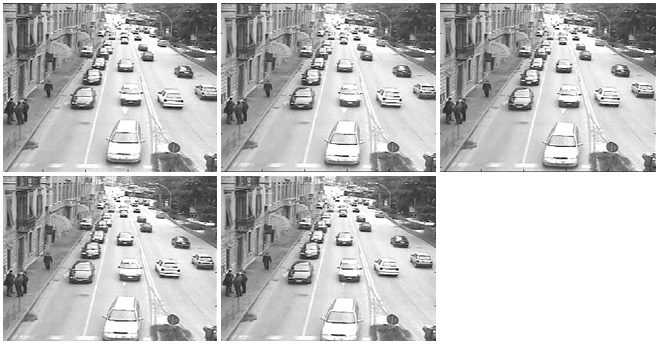}
\end{center}
\caption{The frames that $W_{s}$ contains. The test frame $s$ is the
top-left frame. The frames are ordered from top-left to
bottom-right. }
\label{fig:steady_bg_org}%
\end{figure}

\begin{figure}
\begin{center}
\includegraphics[width=15cm]{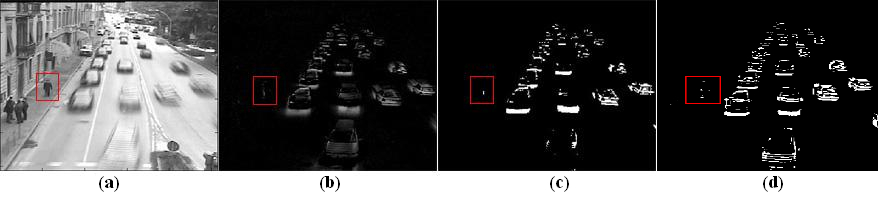}
\end{center}
\caption{(a) The background for the test frame $s$. (b) The test
frame $s$ after the subtraction of the background. (c) The output
for the test frame $s$. (d) The output for the test frame $s$ from
the parallel version of the algorithm. %The person that walks on the
%sidewalk (marked by a square around him), on the left side of the
%frame, is recognized by the SBSDB algorithm.
}
\label{fig:steady_bg}%
\end{figure}

We apply the SBSDB algorithm to a video sequence that consists of
190 grayscale frames of size $256\times 256$. The video sequence was
captured by a static camera and is in AVI format with a frame rate
of 15 fps. The video sequence shows moving cars over a static
background. We apply the sequential version of the algorithm where
the size of the SW is set to 5. We also apply the parallel version
of the algorithm where the video sequence is divided to four blocks
in a $2\times2$ formation. The overlapping size between two (either
horizontally or vertically) adjacent blocks is set to 20 pixels and
the size of the SW is set to 10. Let $s$ be the test frame and let
$W_{s}$ be the SW starting at $s$. In Fig. \ref{fig:steady_bg_org}
we show the frames that $W_{s}$ contains. The output of the SBSDB
algorithm for $s$ is shown in Fig. \ref{fig:steady_bg}.

\subsection{Performance analysis of the DBSDB algorithm}

We apply the DBSDB algorithm to five video sequences. The first four
video sequences are in AVI format with a frame rate of 30 fps. The
last video sequence is in AVI format with a frame rate of 24 fps.
All the video sequences, except the first video sequence, are in RGB
format and are of size $320\times 240$. The first video sequence is
of size $210\times 240$ and is in RGB format. The video sequences
were produced by a static camera and contain a dynamic background.

The input video sequences are:
\begin{enumerate}
\item People walking in front of a fountain.
It contains moving objects in the background such as water flowing,
waving trees and a video screen whose content changes over time. The
DBSDB input is a RTD that contains 170 frames and a BGD that
contains 100 frames. The output of the DBSDB is presented in Fig.
\ref{fig:res_jap}(g).

\begin{figure}
\begin{center}
\includegraphics[width=16cm]{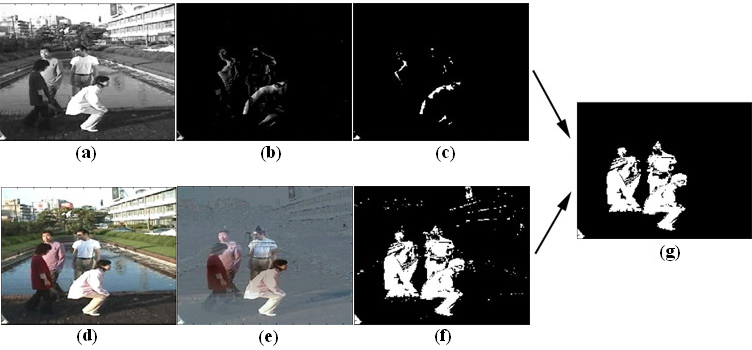}
\end{center}
\caption{(a), (d) The original test frames in grayscale and RGB,
respectively. (b), (e) The grayscale and RGB test frames after the
background subtraction in the classification phase of the DBSDB
algorithm, respectively. (c), (f) Results after the thresholding of
(b) and (e), respectively. (g) The final output of the DBSDB
algorithm after the application of the DFS.}
\label{fig:res_jap}%
\end{figure}

\item A person walking in front of bushes with waving leaves.
The DBSDB input is a RTD that contains 88 frames and a BGD that
contains 160 frames. The output of the DBSDB algorithm is presented
in Fig. \ref{fig:res_bushes}(g).

\begin{figure}
\begin{center}
\includegraphics[width=16cm]{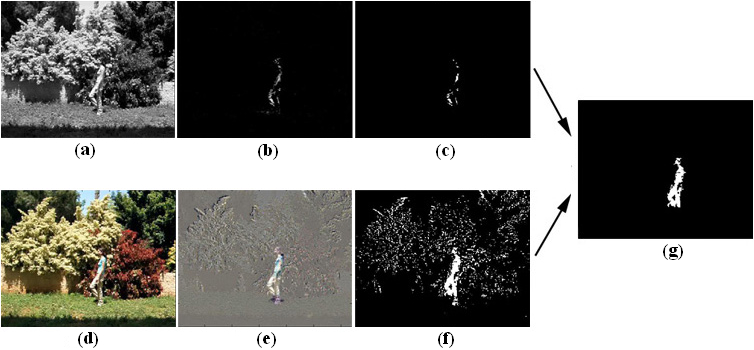}
\end{center}
\caption{(a), (d) The original test frames in grayscale and RGB,
respectively. (b), (e) The grayscale and RGB test frames after the
background subtraction in the classification phase of the DBSDB
algorithm, respectively. (c), (f) Results after the thresholding of
(b) and (e), respectively. (g) The final output of the DBSDB
algorithm after the application of the DFS.}
\label{fig:res_bushes}%
\end{figure}

\item A moving ball in front of waving trees.
The DBSDB input is a RTD that contains 88 frames and a BGD that
contains 160 frames. The output of the DBSDB algorithm is presented
in Fig. \ref{fig:more_res}. Figure \ref{fig:more_res}(d) contains
the result of the sequential version of the algorithm and Fig.
\ref{fig:more_res}(g) contains the results of the parallel version.
In results of the parallel version the video sequence was divided to
four blocks in a $2\times2$ formation. The overlapping size between
two (either horizontally or vertically) adjacent blocks was set to
20 pixels and the size of SW was set to 30.
\item A ball jumping in front of
trees and a car passing behind the trees. The DBSDB input is a RTD
that contains 106 frames and a BGD that contains 160 frames. The
output of the DBSDB algorithm is presented in Fig.
\ref{fig:more_res}(e).
\item A person walking in front of a sprinkler.
The DBSDB input is a RTD that contains 121 frames and a BGD that
contains 100 frames. The output of the DBSDB algorithm is presented
in Fig. \ref{fig:more_res}(f).

\begin{figure}
\begin{center}
\includegraphics[width=15cm]{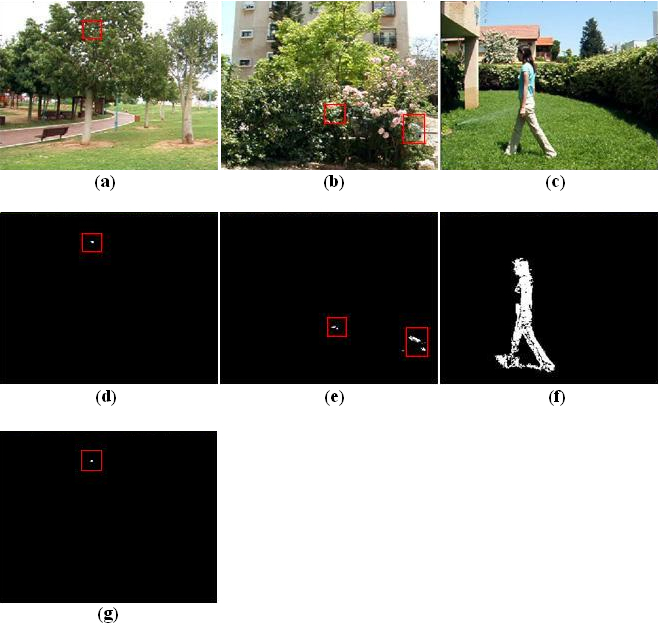}
\end{center}
\caption{ (a)-(c) The original test frames. (d)-(g) The segmented
outputs from the application of the DBSDB algorithm. (a) A ball in
front of waving trees. (d), (g) The result of the sequential and
parallel versions of the algorithm applied on (a), respectively.
(b), (e) A ball jumping in front of a tree and a car passing behind
the trees. (c), (f) A person walking in front of a sprinkler.}
\label{fig:more_res}
\end{figure}

\end{enumerate}

\subsection{Performance comparison between the BSDB algorithm and
other algorithms \label{sub:comparison}}

We compared between the BSDB algorithm and five different background
subtraction algorithms. The input data and the results are taken
from \cite{TKBM99}. All the test sequences were captured by a camera
that has three CCD arrays. The frames are of size 160x120 in RGB
format and are sampled at 4Hz . The test frame that was segmented,
in video sequences where the background changes, is taken to be the
frame that appears 50 frames after the frame where the background
changes. On every output frame (besides the output of the BSDB
algorithm), a speckle removal \cite{TKBM99} was applied to eliminate
islands of 4-connected foreground pixels that contain less than 8
pixels. All other parameters were adjusted for each algorithm in
order to obtain visually optimal results over the entire dataset.
The parameters were used for all sequences. Each test sequence
begins with at least 200 background frames that were used for
training the algorithms, except for the bootstrap sequence. Objects
such as cars, which might be considered foreground in some
applications, were deliberately excluded from the sequences.

Each of the sequences poses a different problem in background
maintenance. The chosen sequences and their corresponding problems
are:

\begin{description}
\item {\textbf{Background object is moved} - }
Problem: A background object can move. These objects should not be
considered as part of the foreground. The sequence contains a person
that walks into a conference room, makes a telephone call, and
leaves with the phone and a chair in a different position. The test
frame is the one that appears 50 frames after the person has left
the scene.
\item {\textbf{Bootstrapping} - }
Problem: A training period without foreground objects is not
available. The sequence contains an overhead view of a cafeteria.
There is constant motion and every frame contains people.
\item {\textbf{Waving Trees} - }
Problem: Backgrounds can contain moving objects. The sequence
contains a person walking in front of a swaying tree.
\item {\textbf{Camouflage} - }
Problem: Pixels of foreground objects may be falsely recognized as
background pixels. The sequence contains a monitor on a desk with
rolling interference bars. A person walks into the scene and stands
in front of the monitor.
\end{description}

We apply six background subtraction algorithms on these sequences,
including the algorithm that is presented in this paper. The
background subtraction algorithms are:
\begin{description}
\item {\textbf{Adjacent Frame Difference} - }
Each frame is subtracted from the previous frame in the sequence.
Absolute differences greater than a threshold are marked as
foreground.
\item {\textbf{Mean and Threshold} - }
Pixel-wise mean values are computed during a training phase, and
pixels within a fixed threshold of the mean are considered
background.
\item {\textbf{Mean and Covariance} - }
The mean and covariance of pixel values are updated continuously
\cite{KWM94}. Foreground pixels are determined by applying a
threshold to the Mahalanobis distance.
\item {\textbf{Mixture of Gaussians} - }
This algorithm is reviewed in section \ref{sub:relatedWork}.
\item {\textbf{Eigen-background} - }
This algorithm is reviewed in section \ref{sub:relatedWork}.
\item {\textbf{BSDB} -}
The algorithm presented in this paper (section \ref{BSDB}).
\end{description}

\begin{figure}
\begin{center}
\includegraphics[width=13cm]{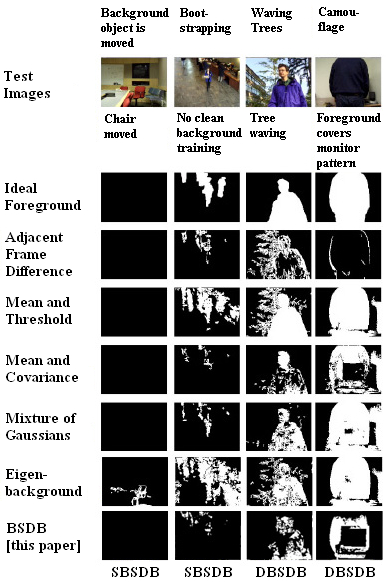}
\end{center}
\caption{The outputs from the applications of the BSDB and five
other algorithms. Each row shows the results of one algorithm, and
each column represents one problem in background maintenance. The
top row shows the test frames. The second row shows the optimal
background outputs.}
\label{fig:comp}%
\end{figure}

The outputs of these algorithms are shown in Fig. \ref{fig:comp}.

We applied the SBSDB algorithm on the first two video sequences: the
moved chair and the bootstrapping. In both cases, the background in
the video sequence is static. In the first video sequence, the SBSDB
algorithm handles the changes in the position of the chair that is a
part of the background. The SBSDB algorithm does not require a
training process so it can handle the second video sequence where
there is no clear background for training. Algorithms that require a
training process can not handle this case.

We applied the DBSDB algorithm on the waving trees and the
camouflage video sequences. In both cases, the background in the
video sequences is dynamic. In the first video sequence, the DBSDB
algorithm captures the movement of the waving trees and eliminates
it from the video sequence. The other algorithms produce false
positive detections. The DBSDB algorithm does not handle well the
last video sequence where the foreground object covers the
background moving object (the monitor). In this case the number of
false negative detections is significant.

\section{Conclusion and Future Work} We introduced in this work the
BSDB algorithm for automatic segmentation of video sequences. The
algorithm contains two versions: the SBSDB algorithm for video
sequences with static background and the DBSDB algorithm for video
sequences that contain dynamic background. The BSDB algorithm
captures the background by reducing the dimensionality of the input
via the DB algorithm. The SBSDB algorithm uses an on-line procedure
while the DBSDB algorithm uses an off-line (training) procedure and
an on-line procedure. During the training phase, the DBSDB algorithm
captures the dynamic background by iteratively applying the DB
algorithm on the background training data. The BSDB algorithm
presents a high quality segmentation of the input video sequences.
Moreover, it was shown that the BSDB algorithm outperforms the
current state-of-the-art algorithms by coping with difficult
situations of background maintenance.

The performance of the BSDB algorithm can be enhanced by improving
the accuracy of the threshold values. Furthermore, it is necessary
to develop a method for automatic computation of $\mu$, which is
used in the threshold computation (sections \ref{sub:GrayThreshold}
and \ref{sub:RGBThreshold}).

Additionally, the output of the BSDB algorithm contains a fair
amount of false negative detections when a foreground object
obscures a brighter background object. This will be improved in
future versions of the algorithm.

The BSDB algorithm can be useful to achieve low-bit rate video
compression for transmission of rich multimedia content. The
captured background is transmitted once followed by the detected
segmented objects.

\bibliographystyle{IEEEtran}
\bibliography{references}

\end{document}